\documentclass[letterpaper]{article} 
\usepackage{aaai25}  
\usepackage{times}  
\usepackage{helvet}  
\usepackage{courier}  
\usepackage[hyphens]{url}  
\usepackage{graphicx} 
\usepackage{natbib}  
\usepackage{caption} 
\usepackage{algorithm}
\usepackage{algorithmic}

\usepackage{newfloat}
\usepackage{listings}

\usepackage{amsmath,epstopdf,amssymb,color,url,multirow,multicol,verbatim,threeparttable,diagbox,makecell,booktabs,color,colortbl}

\newcommand{\etal}{\textit{et al.}}

\definecolor{light-gray}{gray}{0.82}
\DeclareCaptionStyle{ruled}{labelfont=normalfont,labelsep=colon,strut=off} 
\floatstyle{ruled}
\newfloat{listing}{tb}{lst}{}
\floatname{listing}{Listing}
\title{C2P-CLIP: Injecting Category Common Prompt in CLIP to Enhance Generalization in Deepfake Detection}

\author{\small Chuangchuang Tan$^{1,2}$, Renshuai Tao$^{1,2}$, Huan Liu$^{1,2}$, Guanghua Gu$^{3,4}$, Baoyuan Wu$^{5}$, Yao Zhao$^{1,2}$\thanks{Corresponding author},Yunchao Wei$^{1,2}$ \\
{\small $^1$Institute of Information Science, Beijing Jiaotong University}\\
{\small $^2$Beijing Key Laboratory of Advanced Information Science and Network Technology} \\
{\small $^3$School of Information Science and Engineering, Yanshan University}\\
{\small $^4$Hebei Key Laboratory of Information Transmission and Signal Processing}\\
{\small $^5$School of Data Science, The Chinese University of Hong Kong, Shenzhen (CUHK-Shenzhen), China}\\
{\tt\small tanchuangchuang@bjtu.edu.cn}
}

\usepackage{bibentry}

\begin{document}

\maketitle

\begin{abstract}
This work focuses on AIGC detection to develop universal detectors capable of identifying various types of forgery images. 
Recent studies have found large pre-trained models, such as CLIP, are effective for generalizable deepfake detection along with linear classifiers. 
However, two critical issues remain unresolved: 1) understanding why CLIP features are effective on deepfake detection through a linear classifier; and 2) exploring the detection potential of CLIP.  
In this study, we delve into the underlying mechanisms of CLIP's detection capabilities by decoding its detection features into text and performing word frequency analysis. 
Our finding indicates that CLIP detects deepfakes by recognizing similar concepts (Fig. \ref{fig:fig1} a). 
Building on this insight, we introduce Category Common Prompt CLIP, called C2P-CLIP, which integrates the category common prompt into the text encoder to inject category-related concepts into the image encoder, thereby enhancing detection performance  (Fig. \ref{fig:fig1} b). 
Our method achieves a 12.41\% improvement in detection accuracy compared to the original CLIP, without introducing additional parameters during testing. 
Comprehensive experiments conducted on two widely-used datasets, encompassing 20 generation models, validate the efficacy of the proposed method, demonstrating state-of-the-art performance. The code is available at \url{https://github.com/chuangchuangtan/C2P-CLIP-DeepfakeDetection}. 
\end{abstract}

%

\label{Introduction}
\section{Introduction}

\begin{figure}[t]
  \centering
   \includegraphics[scale=0.27]{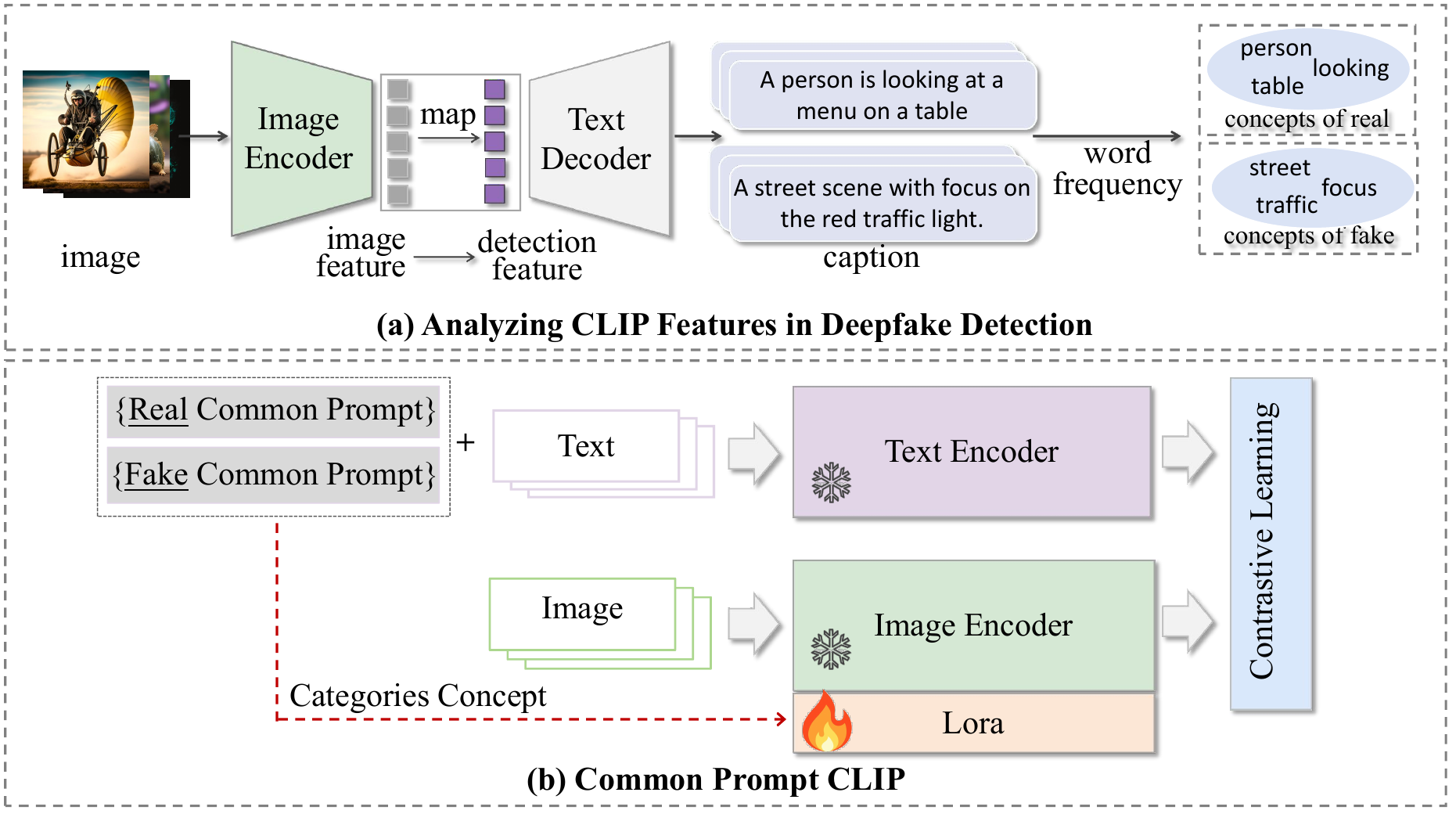}
   \caption{\textbf{Category Common Prompt CLIP.} 
 (a) To investigate the mechanism by which CLIP detects deepfakes, we decode the detection feature into text. Here, the detection features refer to the image features transformed by the linear classifier parameters. 
 Our analysis of these texts reveals that the detection capability arises from the matching of similar concepts. In reality, CLIP does not comprehend ``real" and ``fake", but rather identifies analogous concepts. 
 (b) Building on this insight, we propose a method to enhance the generalization capability of image encoders by introducing a category common prompt. This approach injects manually specified category concepts into the image encoder, aiming to improve its detection performance.
}
   \label{fig:fig1}
\end{figure}

With the development of various image generation techniques, such as Generative Adversarial Networks (GANs)\cite{goodfellow2014generative, karras2018progressive, karras2019style}, diffusion models\cite{ho2020denoising,rombach2022high,zhang2023controlvideo}, etc., distinguishing fake images from real ones has become increasingly challenging for the human eye. As the threshold for image forgery decreases, its misuse will have negative impacts on aspects like the economy and politics. Consequently, the development of universal systems for detecting forged images has become imperative. 

Recently, several forgery detection methods \cite{Frank,li2021frequency} have been developed to identify deepfake images, with a particular focus on detecting face forgery \cite{yan2023ucf, yan2024transcending}. 
Despite these advancements, existing methods often struggle with unseen deepfake sources, leading to inadequate generalization performance. 
To address this issue, various approaches \cite{Tan_2023_CVPR,ojha2023towards,tan2024rethinking} have been proposed to enhance generalization ability. Some studies \cite{Tan_2023_CVPR,tan2024rethinking} focus on developing artifact representations that capture low-level forged traces, such as frequency information, gradients, and neighborhood pixel relationships. 
Other approaches \cite{qian2020thinking,liu2024forgery,tan2024frequency} aim to design more effective detectors to improve overall detection performance.

\begin{figure*}[htb]
  \centering
   \includegraphics[scale=0.72]{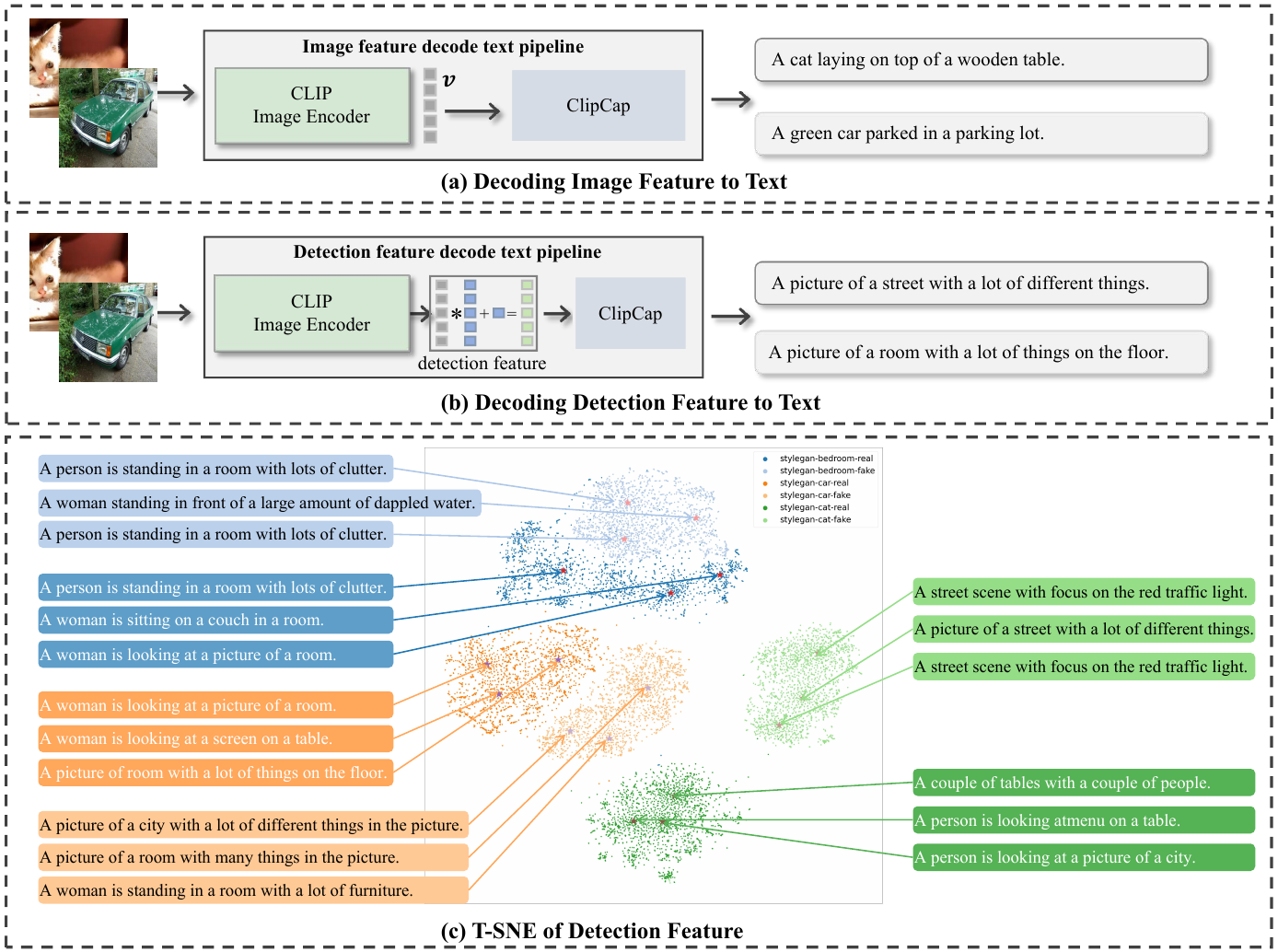}
   \caption{ \textbf{Analyzing CLIP Features in Deepfake Detection.} 
   (a) Decoding Image Feature to Text. We employ ClipCap \cite{mokady2021clipcap} to decode the image feature $v$ to text. 
   (b) Decoding Detection Feature to Text. 
   To discern the specific information within the image features that contribute to classification, we decode the detection features into text.
   The detection features are defined as the combination of image features $v$ and linear classifier $fc$ parameters: $v*fc.weight+fc.bias$. Notice the linear mapping between image features and detection features. Notably, the decoded text bears no direct relevance to the original image content. (c) T-SNE visualization of Detection Feature. 
   We use T-SNE to visualize the detection features from the StyleGAN dataset and decode the textual representations of the three clustering centers within each subset.
   }
   \label{fig:fig2}
\end{figure*}

\begin{figure*}[t]
  \centering
   \includegraphics[scale=0.62]{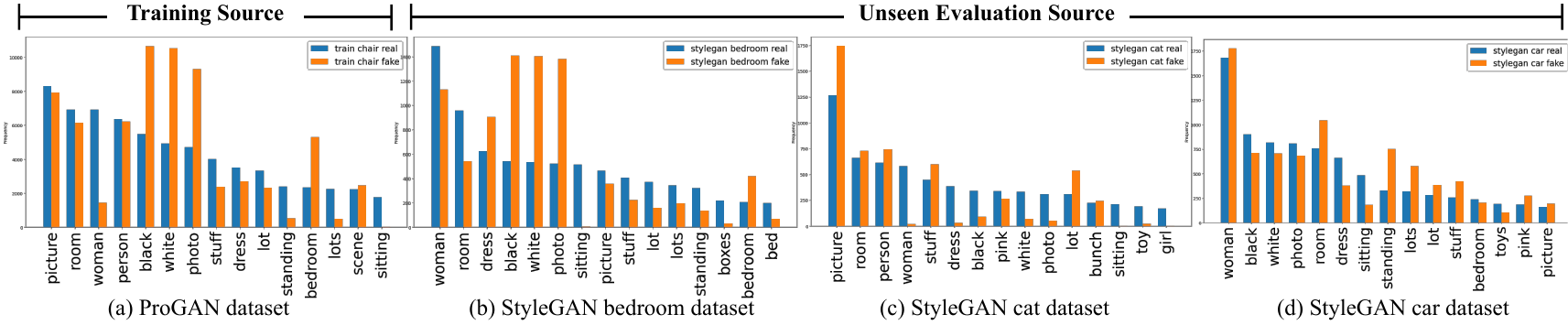}
   \caption{\textbf{Word Frequency Analysis on Various Sources.} We conduct a word frequency analysis on the text decoded from detection features of both the training set (ProGAN) and the unseen test source (StyleGAN). The top 15 words are shown in the graph. The analysis reveals significant differences in word frequencies between the training and test sets. Notably, certain words present in the test set also appear in the training set. For instance, the word 'women' shows substantial frequency variation between (a) and (c). This observation supports the conclusion that CLIP achieves generalizable forgery detection by matching similar concepts or groups of concepts.}
   \label{fig:fig3}
\end{figure*}

Additionally, some research efforts have utilized large pretrained models, such as CLIP\cite{radford2021learning}, to achieve generalizable deepfake detection. 
Notably, UniFd \cite{ojha2023towards} employs the image features extracted by CLIP for linear classification to perform deepfake detection, even with unseen sources. 
FatFormer \cite{liu2024forgery} enhances detection performance by leveraging frequency analysis and using the text encoder as an adaptor for the frozen CLIP vision model. LASTED \cite{wu2023generalizable} focuses on designing textual labels to guide the CLIP vision model through image-text contrastive learning. 
Despite these advancements, two critical issues remain unresolved: 1) elucidating why CLIP, trained with contrastive learning, is capable of achieving generalizable deepfake detection through a linear classifier, and 2) exploring the detection potential of CLIP.

\textbf{Why can CLIP features be used to implement generalization deepfake detection through linear classifiers?} 
In this work, we attempt to discuss this question by decoding CLIP features into text using ClipCap\cite{mokady2021clipcap}, as shown in Figure \ref{fig:fig2}(a)(b). 
Specifically, in UniFd\cite{ojha2023towards}, a linear classifier $fc$ is applied to CLIP's image features $v$ to detect forged images. 
Inspired by this, we define the image features $v$ after transformation by linear classifier $fc$ as detection features. 
To gain insights into how CLIP performs detection, we decode both image features and detection features into text.  
As shown in Figure \ref{fig:fig2}(a), a cat image is processed through the CLIP visual model to extract image features, which are then decoded into text by ClipCap: \underline{A cat laying on top of a wooden table.} However, when we apply the linear classifier from UniFd to obtain detection features and decode them into text, the result is: \underline{A picture of a street with a lot of different things.} 
This observation suggests that semantic information, such as the presence of a cat in the image, is no longer captured in the detection features. We infer that CLIP does not possess inherent true or false semantics; rather, it performs forgery detection by identifying and matching similar concepts.

To validate the above hypothesis, we visualize the detection features of the StyleGAN dataset using T-SNE, as depicted in Figure \ref{fig:fig2}(c). 
We apply k-means clustering to each subset to identify three cluster centers, and the decoded texts for these cluster centers are also shown in Figure \ref{fig:fig2}(c). 
Furthermore, we conduct a word frequency analysis on the text decoded from the detection features of both the training set and the unseen source, with the results illustrated in Figure \ref{fig:fig3}. These findings support our hypothesis that CLIP performs fake detection by identifying similar concepts, rather than by discerning between real and fake images.

\textbf{How to Improve the Detection Performance of CLIP?}  
Building on the aforementioned insight, we propose a novel and effective approach named Category Common Prompt CLIP (C2P-CLIP) to enhance CLIP's detection performance. 
This method utilizes Category Common Prompts to inject category concepts into the image encoder through the text encoder, thereby enhancing its ability to distinguish between real and fake items.  
Specifically, our approach begins by generating text captions for training images using ClipCap. 
We then assign consistent category-specific text prompts to captions from the same category, referred to as category common prompts. For instance, captions of fake samples are paired with the prompt ``Deepfake", while captions of real samples are paired with ``Camera". Subsequently, we retrain CLIP using these new image-text pairs, effectively embedding category-common concepts into the image encoder. During this process, the text encoder's parameters are frozen, and the image encoder is trained using LoRA\cite{hu2021lora}. Overall, we improve the detection capability of CLIP to achieve generalization deepfake detection.

To thoroughly evaluate the generalization capabilities of C2P-CLIP, we conduct extensive simulations using a comprehensive image database generated by 20 distinct models~\footnote{ProGAN, StyleGAN, BigGAN, CycleGAN, StarGAN, GauGAN, Deepfake, SITD, SAN, CRN, IMLE, Guided, LDM, Glide, DALL·E, Midjourney, SDv1.4, SDv1.5, Wukong, VQDM}.
Despite not introducing additional parameters during testing, C2P-CLIP significantly outperforms the original CLIP and achieves state-of-the-art performance.





\section{Related Work}
In this section, we provide a succinct review of the existing literature on deepfake detection methodologies, systematically categorizing them into two principal domains: Face Forgery Detection and AIGC Detection.

\subsection{Face Forgery Detection}
Face forgery detection has been a prominent area of research due to the rise of face edit and generation. 
The domain of face forgery detection has seen substantial advancements, with numerous studies concentrating on the exploitation of spatial or frequency information derived from images.
Rossler \etal \cite{Deepfake} employs the Xception network \cite{chollet2017xception} trained on image data to effectively detect manipulated facial images. Other approaches have focused on detecting specific artifacts in distinct facial regions, such as the eyes \cite{li2018ictu} and lips \cite{haliassos2021lips}. 
In addition, several studies  \cite{luo2021generalizing,masi2020two,qian2020thinking,woo2022add} have explored the use of frequency artifacts to improve the robustness of deepfake detection methodologies. 
To enhance the generalization capability of detection systems, particularly when confronted with unseen sources. A range of methodologies \cite{wang2021representative,chen2022self,cao2022end, he2021beyond, shiohara2022detecting} have been developed to diversify training data, employing strategies such as adversarial training, image reconstruction, and blending techniques. The UCF \cite{yan2023ucf} employs a multi-task learning strategy to extract common forgery features, thereby enhancing the model's generalization capability. Similarly, the LSDA approach \cite{yan2024transcending} constructs and simulates variations within and across forgery features in the latent space, expanding the forgery feature space and enabling the learning of a more generalizable decision boundary.

\subsection{AIGC Detection}
With the advancement of generative technologies, the scope of forged content has expanded beyond facial forgeries to encompass a wide range of scenes. Consequently, recent research has increasingly focused on AIGC detection, which presents unique challenges compared to face forgery detection due to its broader variety of Deepfake types, demanding higher generalization capabilities. 
In this context, CNN-Spot \cite{wang2020cnn} leverages data augmentation techniques to enhance generalization in detection tasks. Several approaches have been proposed to capture low-level artifact representations in AIGC detection, including frequency-based features \cite{jeong2022bihpf}, gradients \cite{Tan_2023_CVPR}, and neighboring pixel relationships \cite{tan2024rethinking}, random-mapping feature\cite{tan2024data}. 
For instance, BiHPF \cite{jeong2022bihpf} amplifies artifact magnitudes through the application of dual high-pass filters, while LGrad \cite{Tan_2023_CVPR} uses gradient information from pre-trained models as artifact representations. NPR \cite{tan2024rethinking} introduces a straightforward yet effective artifact representation by rethinking up-sampling operations. 
In addition to low-level features, large pre-trained models have been employed to capture high-level forging traces for AIGC detection. UniFD \cite{ojha2023towards}, for example, directly utilizes image features from the CLIP model for linear classification, demonstrating effective deepfake detection even with previously unseen sources. FatFormer \cite{liu2024forgery} integrates frequency analysis with a text encoder as an adapter to the frozen CLIP vision model, thereby enhancing detection performance. LASTED \cite{wu2023generalizable} proposes designing textual labels to supervise the CLIP vision model through image-text contrastive learning, further advancing the field of AIGC detection.

\begin{figure*}[t]
  \centering
   \includegraphics[scale=0.63]{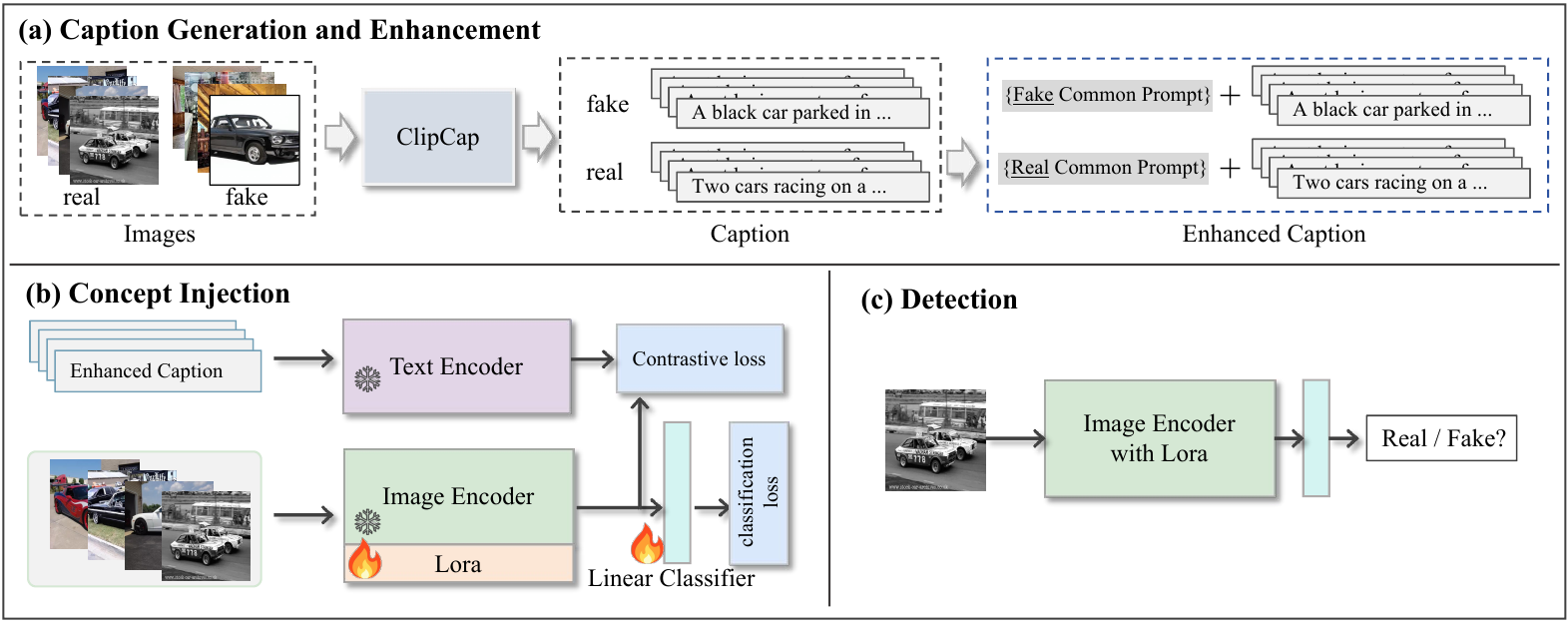}
   \caption{\textbf{Architecture of C2P-CLIP for Generalizable Deepfake Detection.} (a) Caption Generation and Enhancement. We obtain the caption of images using ClipCap, and leverage category common prompts to enhance those text. In this study, we adopt (Trump, Biden), (Deepfake, Camera) as the category common prompts. (b) Concept Injection (Training stage). We use the text-image pair to train the Lora layers and classifier by contrastive loss and classification loss. (c) Detection (Testing stage). Only image encoder and classifier are utilized to perform detection. }
   \label{fig:overall}
\end{figure*}

\section{Methodology}
In this section, we introduce our C2P-CLIP, a universal approach designed for generalizable deepfake detection. The overall architecture is illustrated in Figure \ref{fig:overall}.

\subsection{Overall Architecture} 
In this work, we focus on enhancing the detection performance of CLIP in a simple and effective manner. 
To achieve this, we first investigate why CLIP features can be generalized in deepfake detection through linear classifiers, as discussed in Section 2. 
Building on the analysis, we propose a simple yet effective method called Category Common Prompt CLIP (C2P-CLIP) to achieve generalizable deepfake detection. The comprehensive architecture of the C2P-CLIP approach is illustrated in Figure \ref{fig:overall}.
Our method comprises three key components:
\begin{itemize}
\item \textbf{Caption Generation and Enhancement}: We begin by generating captions for images and enhancing them by appending a Category Common Prompt.
\item \textbf{Concept Injection}. Next, we incorporate classification concepts from the enhanced captions into the image encoder using a combination of contrastive loss and classification loss.
\item \textbf{Detection}: Finally, we integrate LoRA parameters into the image encoder to facilitate detection without adding additional parameters during testing.
\end{itemize}

\subsection{Caption Generation and Enhancement}
The ability of CLIP features to perform forgery detection through linear classification is a fascinating phenomenon. As discussed in Section 2, we attribute this capability to CLIP's mechanism of seeking similar concepts. Building on this insight, we hypothesize that injecting explicit classification concepts into the image encoder can significantly enhance detection performance. 
To this end, the category common prompts are designed to enhance the captions associated with the images. These enhanced captions are then utilized in contrastive learning to embed the classification concepts into the image encoder. Specifically, during the training stage, we append a consistent prompt, such as ``Camera" or  ``Biden", to all captions of real images, and a different prompt, such as ``Deepfake" or ``Trump", to the captions of fake images. 

Let us consider a training dataset $X$ containing both real and fake images, defined as follows:
\begin{equation}
\begin{split}
    X =  \{ x_{j}, y_{j}\}_{j=1}^{N}, \quad y \in \{0, 1\},
 \end{split}
  \label{eq:eq1}
\end{equation}
where $y=1$ indicates that the image is fake, and $y=0$ indicates that the image is real. 
For each image in the training set, we obtain its corresponding caption using the ClipCap model. The set of captions associated with the training images is denoted as $C$, defined as:
\begin{equation}
\begin{split}
    &C = \{ c_{j}, y_{j}\}_{j=1}^{N}, \quad y \in\{0,1\}.
 \end{split}
  \label{eq:eq2}
\end{equation}

Next, we utilize the category common prompts $P =\{P_{real}, P_{fake}\}$ to enhance the captions as follows:
\begin{equation}
\begin{split}
& \widetilde{C} = \{ \widetilde{c_{j}} \}_{j=1}^{N}, \\
& \widetilde{c_{j}} = \left\{
\begin{aligned}
(P_{real}, c_j) ,& \quad \text{if } y=0, \\
(P_{fake}, c_j) ,& \quad \text{if } y=1,
\end{aligned}
\right.
\end{split}
\label{eq:eq3}
\end{equation}
where $P =\{P_{real}, P_{fake}\}$ are typically assigned as pairs of words different from the image caption, such as $(P_{real}=Camera, P_{fake}=Deepfake)$ or $(P_{real}=Biden, P_{fake}=Trump)$. 
These category common prompts, when appended to the original captions, enhance the textual context, enabling the model to better distinguish between real and fake images. The enhanced captions $\widetilde{C}$ are then used in contrastive learning to transmit the information from the category common prompts into the image encoder.

\subsection{Concept Injection (Training Stage)}
To embed the deepfake detection concept from the enhanced captions into the image encoder, we employ contrastive learning during the training stage. In this phase, the image- and text encoder are kept frozen, while the Lora layers are applied to the image encoder. The goal is to transfer the categorical concepts embedded in the enhanced captions into the image encoder, thereby improving its deepfake detection capabilities. To achieve this, we use two key losses: contrastive loss and classification loss. Together, these losses guide the image encoder to effectively learn and apply the deepfake detection concepts during training, enhancing its generalization performance across various unseen sources.

Specifically, we first compute text features $u$ and image features $v$ as follows::
\begin{equation}
\begin{split}
    &u_j = encoder_{text}(\widetilde{c_j}), \\
    &v_j = encoder_{img}^{lora}(x_j).
 \end{split}
  \label{eq:eq4}
\end{equation}
Here, $encoder_{text}()$ denotes the text encoder, and $encoder_{img}^{lora}()$ refers to the image encoder equipped with the Lora layer.  
Next, we compute the contrastive learning loss as follows:
\begin{equation}
\begin{split}
    &\mathcal L_{contrastive} = (\mathcal L_{v->u}+\mathcal L_{u->v})/2, \\
    & \mathcal L_{v->u} = -\frac{1}{N} \sum_{i}^N log \frac{exp(v_i^T u_i)}{\sum_{j=1}^N exp(v_i^T u_j)}, \\
    & \mathcal L_{u->v} = -\frac{1}{N} \sum_{i}^N log \frac{exp(u_i^T v_i)}{\sum_{j=1}^N exp(u_i^T v_j)}.
 \end{split}
  \label{eq:eq5}
\end{equation}

In addition, we also apply a linear classifier $Linear$ on image features $v$ to perform the classification. The cross-entropy loss is adopted as the classification loss $\mathcal L_{classification}$.
The final loss function is obtained by the weighted sum of the above loss functions.
\begin{equation}
\begin{split}
    &\mathcal L = \mathcal L_{contrastive} + \alpha * \mathcal L_{classification}, 
 \end{split}
  \label{eq:eq6}
\end{equation}
where $\alpha$ are hyper-parameters for balancing two losses.

\subsection{Detection (Testing Stage)}
During the evaluation phase, only the image encoder and the classifier are utilized to perform the detection. The detection process is as follows:
\begin{equation}
\begin{split}
    &p = classifier( \widetilde{encoder_{img}}(x) ).
 \end{split}
  \label{eq:eq7}
\end{equation}
Here, $\widetilde{encoder_{img}}$ represents the image encoder integrated with the Lora parameters, which have been fine-tuned during the training stage. 


\begin{table*}[!ht]
    \centering
\resizebox{\textwidth}{20mm}{
    \begin{tabular}{l c c c c c c c c c c c c c c c c c c c c c}
    \hline
      \multirow{3}*{Methods}  & \multirow{3}*{Ref} & \multicolumn{6}{c}{GAN} &  \multirow{3}*{\makecell[c]{Deep\\fakes}} & \multicolumn{2}{c}{Low level} & \multicolumn{2}{c}{Perceptual loss} &  \multirow{3}*{Guided} & \multicolumn{3}{c}{LDM} & \multicolumn{3}{c}{Glide} &  \multirow{3}*{Dalle}  &  \multirow{3}*{mAP}\\ 
      
       \cmidrule(r){3-8} \cmidrule(r){10-11}  \cmidrule(r){12-13}  \cmidrule(r){15-17}  \cmidrule(r){18-20}  
       ~ & ~ & \makecell[c]{Pro-\\GAN} & \makecell[c]{Cycle-\\GAN} & \makecell[c]{Big-\\GAN} & \makecell[c]{Style-\\GAN}  & \makecell[c]{Gau-\\GAN}  & \makecell[c]{Star-\\GAN} & ~ & \makecell[c]{\\SITD}& \makecell[c]{\\SAN}& \makecell[c]{\\CRN}& \makecell[c]{\\IMLE}& ~ & {\makecell[c]{200\\steps}}& {\makecell[c]{200\\w/cfg}}& {\makecell[c]{100\\steps}}& {\makecell[c]{100\\27}} & {\makecell[c]{50\\27}} & \makecell[c]{100\\10} & ~ & ~\\
       \hline
       {CNN-Spot}  & CVPR2020 & 100.0 & 93.47 & 84.5 & 99.54 & 89.49 & 98.15 & 89.02 & 73.75 & 59.47 & 98.24 & 98.4 & 73.72 & 70.62 & 71.0 & 70.54 & 80.65 & 84.91 & 82.07 & 70.59 & 83.58 \\
       {Patchfor} & ECCV2020 & 80.88 & 72.84 & 71.66 & 85.75 & 65.99 & 69.25 & 76.55 & 76.19 & 76.34 & 74.52 & 68.52 & 75.03 & 87.1 & 86.72 & 86.4 & 85.37 & 83.73 & 78.38 & 75.67 & 77.73 \\
       Co-occurence & Elect. Imag. & 99.74 & 80.95 & 50.61 & 98.63 & 53.11 & 67.99 & 59.14 & 68.98 & 60.42 & 73.06 & 87.21 & 70.20 & 91.21 & 89.02 & 92.39 & 89.32 & 88.35 & 82.79 & 80.96 & 78.11 \\
       Freq-spec & WIFS2019 & 55.39 & 100.0 & 75.08 & 55.11 & 66.08 & 100.0 & 45.18 & 47.46 & 57.12 & 53.61 & 50.98 & 57.72 & 77.72 & 77.25 & 76.47 & 68.58 & 64.58 & 61.92 & 67.77 & 66.21 \\
       F3Net  & ECCV2020  & 99.96 & 84.32 & 69.90 & 99.72 & 56.71 & 100.0 & 78.82 & 52.89 & 46.70 & 63.39 & 64.37 & 70.53 & 73.76 & 81.66 & 74.62 & 89.81 & 91.04 & 90.86 & 71.84 & 76.89  \\
       UniFD  & CVPR2023 & 100.0 & 98.13 & 94.46 & 86.66 & 99.25 & 99.53 & 91.67 & 78.54 & 67.54 & 83.12 & 91.06 & 79.24 & 95.81 & 79.77 & 95.93 & 93.93 & 95.12 & 94.59 & 88.45 & 90.14 \\
       LGrad  & CVPR2023  &100.0 & 93.98 & 90.69 & 99.86 & 79.36 & 99.98 & 67.91 & 59.42 & 51.42 & 63.52 & 69.61 & 87.06 & 99.03 & 99.16 & 99.18 & 93.23 & 95.10 & 94.93 & 97.23 & 86.35  \\
       FreqNet & AAAI2024  & 99.92 & 99.63 & 96.05 & 99.89 & 99.71 & 98.63 & 99.92 & 94.42 & 74.59 & 80.10 & 75.70 & 96.27 & 96.06 & 100.0 & 62.34 & 99.80 & 99.78 & 96.39 & 77.78 & 91.95 \\ 
       NPR  & CVPR2024  &  100.0 & 99.53 & 94.53 & 99.94 & 88.82 & 100.0 & 84.41 & 97.95 & 99.99 & 50.16 & 50.16 & 98.26 & 99.92 & 99.91 & 99.92 & 99.87 & 99.89 & 99.92 & 99.26 & 92.76  \\
       FatFormer & CVPR2024  & 100.0 & 100.0 & 99.98 & 99.75 & 100.0 & 100.0 & 97.99 & 97.94 & 81.21 & 99.84 & 99.93 & 91.99 & 99.81 & 99.09 & 99.87 & 99.13 & 99.41 & 99.20 & 99.82 & \underline{98.16} \\
      \hline
       \rowcolor{light-gray} Ours  & Trump,Biden & 99.99 & 99.88 & 99.87 & 99.98 & 99.96 & 100.0 & 97.28 & 99.87 & 76.00 & 99.78 & 99.93 & 92.23 & 99.98 & 99.79 & 99.98 & 99.30 & 99.32 & 99.38 & 99.94 & 98.02 \\
       \rowcolor{light-gray} Ours & {\scriptsize Deepfake,Camera } & 100.0 & 100.0 & 99.96 & 99.50 & 100.0 & 100.0 & 98.59 & 98.92 & 84.56 & 99.86 & 99.95 & 94.13 & 99.99 & 99.83 & 99.98 & 99.72 & 99.79 & 99.83 & 99.91 & \textbf{98.66} \\
\hline
    \end{tabular}
}
  \caption{\textbf{Cross-model Average Precision (AP) Performance on the UniversalFakeDetect Dataset.}  We copy the results of CNN-Spot\shortcite{wang2020cnn}, Patchfor\shortcite{chai2020makes}, Co-occurence\shortcite{nataraj2019detecting}, Freq-spec\shortcite{zhang2019detecting}, and UniFD\shortcite{ojha2023towards} from paper \cite{ojha2023towards}, and obtain results of F3Net\shortcite{qian2020thinking}, LGrad\shortcite{Tan_2023_CVPR}, FreqNet\shortcite{tan2024frequency}, NPR\shortcite{tan2024rethinking}, and FatFormer\shortcite{liu2024forgery} using the official pre-trained model or re-implemented model. \textbf{Bold} and \underline{underline} represent the best and second-best performance, respectively. }
  \label{tab:SOTA1}
\end{table*}

\begin{table*}[!ht]
    \centering
\resizebox{\textwidth}{20mm}{
    \begin{tabular}{l c c c c c c c c c c c c c c c c c c c c c}
    \hline
      \multirow{3}*{Methods}  & \multirow{3}*{Ref} & \multicolumn{6}{c}{GAN} &  \multirow{3}*{\makecell[c]{Deep\\fakes}} & \multicolumn{2}{c}{Low level} & \multicolumn{2}{c}{Perceptual loss} &  \multirow{3}*{Guided} & \multicolumn{3}{c}{LDM} & \multicolumn{3}{c}{Glide} &  \multirow{3}*{Dalle}  &  \multirow{3}*{mAcc}\\ 
      
       \cmidrule(r){3-8} \cmidrule(r){10-11}  \cmidrule(r){12-13}  \cmidrule(r){15-17}  \cmidrule(r){18-20}  
       ~ & ~ & \makecell[c]{Pro-\\GAN} & \makecell[c]{Cycle-\\GAN} & \makecell[c]{Big-\\GAN} & \makecell[c]{Style-\\GAN}  & \makecell[c]{Gau-\\GAN}  & \makecell[c]{Star-\\GAN} & ~ & \makecell[c]{\\SITD}& \makecell[c]{\\SAN}& \makecell[c]{\\CRN}& \makecell[c]{\\IMLE}& ~ & {\makecell[c]{200\\steps}}& {\makecell[c]{200\\w/cfg}}& {\makecell[c]{100\\steps}}& {\makecell[c]{100\\27}} & {\makecell[c]{50\\27}} & \makecell[c]{100\\10} & ~ & ~\\
       \hline
       CNN-Spot & CVPR2020 & 99.99 & 85.20 & 70.20 & 85.7 & 78.95 & 91.7 & 53.47 & 66.67 & 48.69 & 86.31 & 86.26 & 60.07 & 54.03 & 54.96 & 54.14 & 60.78 & 63.8 & 65.66 & 55.58 & 69.58 \\
       Patchfor & ECCV2020 & 75.03 & 68.97 & 68.47 & 79.16 & 64.23 & 63.94 & 75.54 & 75.14 & 75.28 & 72.33 & 55.3 & 67.41 & 76.5 & 76.1 & 75.77 & 74.81 & 73.28 & 68.52 & 67.91 & 71.24 \\
       Co-occurence & Elect. Imag. &  97.70 & 63.15 & 53.75 & 92.50 & 51.1 & 54.7 & 57.1 & 63.06 & 55.85 & 65.65 & 65.80 & 60.50 & 70.7 & 70.55 & 71.00 & 70.25 & 69.60 & 69.90 & 67.55 & 66.86 \\
       Freq-spec  & WIFS2019 & 49.90 & 99.90 & 50.50 & 49.90 & 50.30 & 99.70 & 50.10 & 50.00 & 48.00 & 50.60 & 50.10 & 50.90 & 50.40 & 50.40 & 50.30 & 51.70 & 51.40 & 50.40 & 50.00 & 55.45 \\
       F3Net & ECCV2020 & 99.38 & 76.38 & 65.33 & 92.56 & 58.10 & 100.0 & 63.48 & 54.17 & 47.26 & 51.47 & 51.47 & 69.20 & 68.15 & 75.35 & 68.80 & 81.65 & 83.25 & 83.05 & 66.30 & 71.33  \\
       UniFD & CVPR2023 &100.0 & 98.50 & 94.50 & 82.00 & 99.50 & 97.00 & 66.60 & 63.00 & 57.50 & 59.5 & 72.00 & 70.03 & 94.19 & 73.76 & 94.36 & 79.07 & 79.85 & 78.14 & 86.78 & 81.38 \\
       LGrad & CVPR2023 & 99.84 & 85.39 & 82.88 & 94.83 & 72.45 & 99.62 & 58.00 & 62.50 & 50.00 & 50.74 & 50.78 & 77.50 & 94.20 & 95.85 & 94.80 & 87.40 & 90.70 & 89.55 & 88.35 & 80.28  \\
       FreqNet & AAAI2024  & 97.90 & 95.84 & 90.45 & 97.55 & 90.24 & 93.41 & 97.40 & 88.92 & 59.04 & 71.92 & 67.35 & 86.70 & 84.55 & 99.58 & 65.56 & 85.69 & 97.40 & 88.15 & 59.06 & 85.09 \\ 
       NPR & CVPR2024 & 99.84 & 95.00 & 87.55 & 96.23 & 86.57 & 99.75 & 76.89 & 66.94 & 98.63 & 50.00 & 50.00 & 84.55 & 97.65 & 98.00 & 98.20 & 96.25 & 97.15 & 97.35 & 87.15 & 87.56 \\
       FatFormer & CVPR2024  & 99.89 & 99.32 & 99.50 & 97.15 & 99.41 & 99.75 & 93.23 & 81.11 & 68.04 & 69.45 & 69.45 & 76.00 & 98.60 & 94.90 & 98.65 & 94.35 & 94.65 & 94.20 & 98.75 & 90.86 \\
       \hline
       \rowcolor{light-gray} Ours & Trump,Biden & 99.71 & 90.69 & 95.28 & 99.38 & 95.26 & 96.60 & 89.86 & 98.33 & 64.61 & 90.69 & 90.69 & 77.80 & 99.05 & 98.05 & 98.95 & 94.65 & 94.20 & 94.40 & 98.80 & \underline{93.00} \\
       \rowcolor{light-gray} Ours & {\scriptsize Deepfake,Camera} & 99.98 & 97.31 & 99.12 & 96.44 & 99.17 & 99.60 & 93.77 & 95.56 & 64.38 & 93.29 & 93.29 & 69.10 & 99.25 & 97.25 & 99.30 & 95.25 & 95.25 & 96.10 & 98.55 & \textbf{93.79} \\
\hline
    \end{tabular}
}
  \caption{\textbf{Cross-model Accuracy (Acc) Performance on the UniversalFakeDetect Dataset.} 
  }
  \label{tab:SOTA2}
\end{table*}


\section{Experiments}
In this section, we provide the evaluation encompasses various aspects such as datasets, implementation details, and deepfake detection performance.

\label{ER}
\subsection{Datasets}
To demonstrate the detection performance of our C2P-CLIP, we conducted the generalization evaluation on two widely used datasets following baselines\cite{ojha2023towards,zhu2024genimage}, including 
UniversalFakeDetect dataset\cite{ojha2023towards} and GenImage dataset\cite{zhu2024genimage}.

\subsubsection{UniversalFakeDetect Dateset}
This dataset uses ProGAN as the training set following \cite{wang2020cnn}, which includes 20 subsets of generated images. For training, we adopt the 4-class setting $(horse, chair, cat, car)$ following \cite{tan2024rethinking,liu2024forgery}.
The test set comprises 19 subsets from various generative models, including: 
ProGAN \cite{karras2018progressive}, StyleGAN \cite{karras2019style}, BigGAN \cite{BigGAN}, CycleGAN \cite{CycleGAN}, StarGAN \cite{choi2018stargan}, GauGAN \cite{GauGAN} and Deepfake \cite{Deepfake}, CRN \cite{chen2017photographic}, IMLE \cite{li2019diverse}, SAN \cite{dai2019second}, SITD \cite{chen2018learning}, Guided diffusion model\cite{dhariwal2021diffusion}, LDM \cite{rombach2022high}, Glide \cite{nichol2021glide}, DALLE \cite{ramesh2021zero}.

\subsubsection{Genimage dataset} This dataset primarily employs the Diffusion model for image generation, including Midjourney\cite{Midjourney}, SDv1.4\cite{rombach2022high}, SDv1.5\cite{rombach2022high}, ADM\cite{dhariwal2021diffusion}, GLIDE\cite{nichol2021glide}, Wukong\cite{Wukong}, VQDM\cite{gu2022vector}, BigGAN\cite{BigGAN}. Following the settings on GenImage, we use SDv1.4 as the training set and the remaining models as the test set.

\begin{table*}[!ht]
    \centering
\resizebox{0.8\textwidth}{22mm}{
    \begin{tabular}{l c c c c c c c c c | c}
    \bottomrule \hline
        Method & Ref & Midjourney & SDv1.4 & SDv1.5 & ADM & GLIDE & Wukong & VQDM & BigGAN & mAcc\\
          \bottomrule \hline
ResNet-50\shortcite{he2016deep}  &  CVPR2016   & 54.9 & 99.9 & 99.7 & 53.5 & 61.9 & 98.2 & 56.6 & 52.0 & 72.1 \\
DeiT-S\shortcite{touvron2021training}& ICML2021 & 55.6 & 99.9 & 99.8 & 49.8 & 58.1 & 98.9 & 56.9 & 53.5 & 71.6 \\
Swin-T\shortcite{liu2021swin}    &  ICCV2021   & 62.1 & 99.9 & 99.8 & 49.8 & 67.6 & 99.1 & 62.3 & 57.6 & 74.8 \\
CNNSpot\shortcite{wang2020cnn}    &  CVPR2020  & 52.8 & 96.3 & 95.9 & 50.1 & 39.8 & 78.6 & 53.4 & 46.8 & 64.2 \\
Spec\shortcite{zhang2019detecting} & WIFS2019  & 52.0 & 99.4 & 99.2 & 49.7 & 49.8 & 94.8 & 55.6 & 49.8 & 68.8 \\
F3Net\shortcite{qian2020thinking}  & ECCV2020  & 50.1 & 99.9 & 99.9 & 49.9 & 50.0 & 99.9 & 49.9 & 49.9 & 68.7 \\
GramNet\shortcite{liu2020global}   & CVPR2020  & 54.2 & 99.2 & 99.1 & 50.3 & 54.6 & 98.9 & 50.8 & 51.7 & 69.9 \\
UnivFD\shortcite{ojha2023towards}  & CVPR2023  & 93.9 & 96.4 & 96.2 & 71.9 & 85.4 & 94.3 & 81.6 & 90.5 & 88.8 \\
NPR \shortcite{tan2024rethinking}  &  CVPR2024 & 81.0 & 98.2 & 97.9 & 76.9 & 89.8 & 96.9 & 84.1 & 84.2 & 88.6 \\
FreqNet \shortcite{tan2024frequency} & AAAI2024 & 89.6 & 98.8 & 98.6 & 66.8 & 86.5 & 97.3 & 75.8 & 81.4 & 86.8 \\
FatFormer\shortcite{liu2024forgery} & CVPR2024 & 92.7 & 100.0 & 99.9 & 75.9 & 88.0 & 99.9 & 98.8 & 55.8 & 88.9 \\
\hline
\rowcolor{light-gray} Ours& Trump,Biden &  82.2 & 95.1 & 95.5 & 95.1 & 98.9 & 98.7 & 93.8 & 98.3 & \underline{94.7}  \\
\rowcolor{light-gray} Ours & Deepfake,Camera & 88.2 & 90.9 & 97.9 & 96.4 & 99.0 & 98.8 & 96.5 & 98.7 & \textbf{95.8} \\
\bottomrule
    \end{tabular}
}
  \caption{\textbf{Cross-model Accuracy (Acc) Performance on the Genimage Dataset.} The SDv1.4 is employed as the training set following \cite{zhu2024genimage}.  We copy the results of ResNet-50, DeiT-S, Swin-T, CNNSpot, Spec, F3Net, GramNet from paper \cite{zhu2024genimage}, and obtain the results of UnivFD, FreqNet, NPR, and FatFormer using re-implement model.}
  \label{tab:SOTA3}
\end{table*}

\begin{figure*}[t]
  \centering
   \includegraphics[scale=0.60]{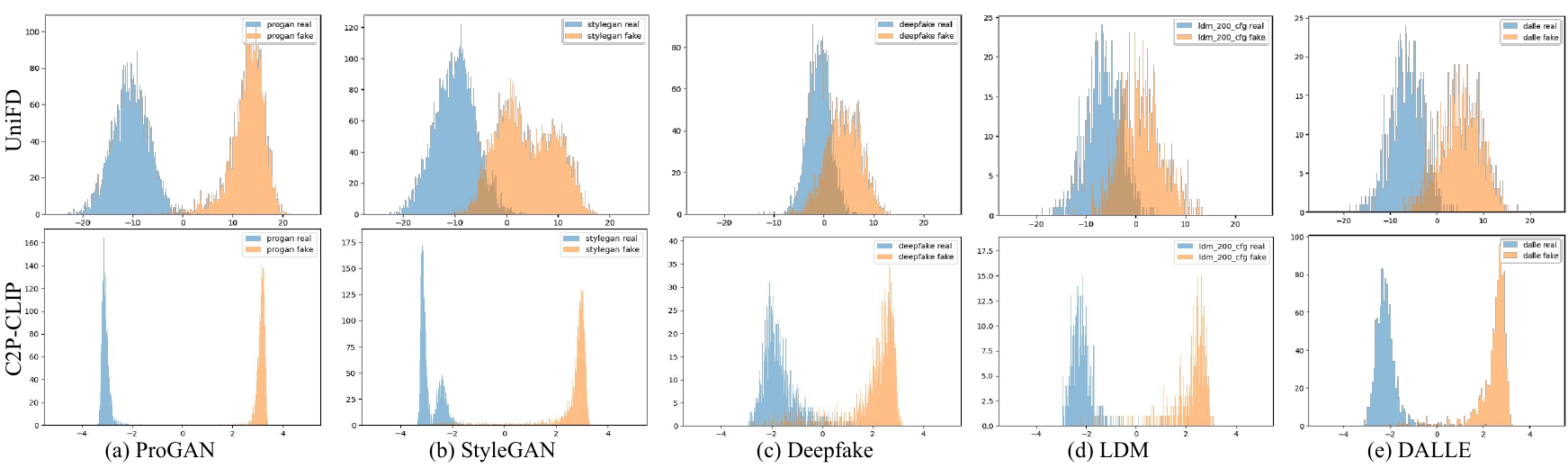}
   \caption{\textbf{Logit distributions of extracted forgery features.} We compare the baseline UniFD and our C2P-CLIP. A total of four testing GANs and diffusion models are considered, including ProGAN, StyleGAN, Deepfake, LDM, and DALLE.}
   \label{fig:fig5}
\end{figure*}

\subsubsection{Implementation Details}

We utilize the Adam optimizer \cite{kingma2015adam} with an initial learning rate of \(4 \times 10^{-4}\). The batch size is set to 128, and we train the model for 1 epoch. The ViT-L/14 model of CLIP is adopted as the pre-trained model following the baseline UniFD\cite{ojha2023towards}. We apply Lora layers on the $q\_{proj}$, $k\_{proj}$, and $v\_{proj}$ layers using the Parameter-Efficient Fine-Tuning (PEFT) \cite{peft} library. For the setting of the Lora layers, we configure the hyperparameters as follows: $lora\_r = 6$, $lora\_alpha = 6$, and $lora\_dropout = 0.8$. 
The hyperparameter $\alpha$ is set to 8.0. A random seed of 123 is used for reproducibility. The proposed method is implemented using Pytorch\cite{paszke2019pytorch} on 4 Nvidia GeForce RTX 4090 GPUs.


Following the baselines \cite{ojha2023towards, liu2024forgery}, we use mean average precision (mAP) and mean accuracy (mAcc) as evaluation metrics.

\subsection{Quantitative analysis}
Here we demonstrate the cross-model detection capability of the proposed method on two datasets. 

\textbf{Evaluation on UniversalFakeDetect}
The results of average precision (AP) and accuracy (Acc) are shown in Table \ref{tab:SOTA1} and \ref{tab:SOTA2}, respectively. The UniversalFakeDetect dataset consists of 19 generators, including GANs, Deepfake, low-level vision models, perceptual loss models, and diffusion models. Our method is trained using ProGAN with 4 training settings, achieving detection results of 93.79\% accuracy and 98.66\% mAP on the 19 test subsets. 
The baseline UniFD directly uses the original CLIP for deepfake detection. In contrast, our method injects category concepts into the visual encoder. Compared to UniFD, our method improves mAcc by 12.41\% and mAP by 8.52\%. This indicates that the proposed category common prompts effectively enhance CLIP's deepfake detection capability.
Furthermore, compared to the latest state-of-the-art method, FatFormer, our method improves accuracy by 2.93\%. FatFormer employs frequency blocks and text encoders as adapters to fine-tune the visual encoder, which adds extra inference parameters. In contrast, our method only uses text encoders and Lora parameters during training, without adding parameters during testing, thus achieving a simple but effective improvement in CLIP's detection performance.

Additionally, we test the detection performance with different categories common prompts. Using prompts (Trump, Biden) and (Deepfake, Camera), we achieved detection accuracies of 93.79\% and 93.00\%, respectively. This demonstrates that our method can improve detection performance without relying on specific prompts.

\textbf{Evaluation on Genimage}
We show the results on the Genimage dataset in Table \ref{tab:SOTA3}. The results of accuracy (Acc) are presented. The Genimage dataset includes 7 diffusion models and one GAN model, focusing primarily on the detection performance of methods on recent diffusion models. Due to the varying image sizes in the Genimage dataset, images smaller than 224 pixels are duplicated and then cropped to 224 pixels. We utilize the same setting to re-implement UnivFD, FreqNet, FatFormer, and NPR.
When using SDv1.4 as the training set, our method achieves a 95.8\% accuracy rate on the test set. Compared with the baseline UniFD, our method improves accuracy by 7.0\%. Furthermore, compared to the state-of-the-art methods, the proposed method improved accuracy by {6.9\%}. This demonstrates that our method performs well with diffusion models for training. 
Additionally, we also train on the Genimage dataset using different categories common prompts. When using prompts (Trump, Biden) and (Deepfake, Camera), the accuracy rates reached 94.7\% and 95.8\%, respectively.

\subsection{Qualitative Analysis}
To further assess the generalization capability of our method, we visualize the logit distributions of both UniFD and our approach, as illustrated in Figure\ref{fig:fig5}. 
This visualization sheds light on the degree of separation between 'real' and 'fake' images during the evaluation phase, thus indicating the effectiveness of our method in generalizing across various forgery representations.
Our analysis shows the considerable overlap between 'real' and 'fake' regions when the baseline UniFD encounters previously unseen GANs or diffusion models, leading to the misclassification of forgeries as 'real'. In contrast, our approach maintains clear differentiation between 'real' and 'fake' categories, even when confronted with unseen sources. These findings underscore the robustness of our method in enhancing the separation between 'real' and 'fake' classes and demonstrate its superior generalization across a diverse array of image sources.



\section{Conclusion}
In this study, we first sought to understand why CLIP features are effective for deepfake detection through a linear classifier. To investigate what information in the image features contributes to the detection, we decode the detection features into text and conduct a word frequency analysis. As far as we know, this is the first time this issue has been approached from this perspective. We conclude that CLIP achieves classification by matching similar concepts rather than discerning true and false.
Based on this conclusion, we propose category common prompts to fine-tune the image encoder by manually constructing category concepts combined with contrastive learning. This approach led to an improvement in detection performance.
However, a limitation of our method is that, while it uses word frequency to analyze detection features, it lacks a comprehensive analysis of the entire caption, resulting in incomplete information. We aim to address this issue in future work.

\bigskip

\bibliography{aaai25}

\end{document}